\documentclass{article}
\usepackage{ijcai24}

\usepackage{times}
\usepackage{soul}
\usepackage{url}
\usepackage[hidelinks]{hyperref}
\usepackage[utf8]{inputenc}
\usepackage[small]{caption}
\usepackage{graphicx}
\usepackage{amsmath}
\usepackage{amsthm}
\usepackage{booktabs}
\usepackage{algorithm}
\usepackage{algorithmic}
\usepackage[switch]{lineno}
\usepackage{comment}

\usepackage{amsmath}
\usepackage{amssymb}
\usepackage{booktabs}
\usepackage{times}
\usepackage{epsfig}
\usepackage{enumitem}
\usepackage{pifont}

\usepackage{wrapfig}
\usepackage{caption}
\usepackage{subcaption}
\usepackage{amsthm}
\usepackage{amssymb}
\usepackage{tabularx}
\usepackage{enumitem}
\usepackage{multirow}
\usepackage{makecell}
\usepackage{wrapfig}
\usepackage[table, dvipsnames]{xcolor} 
\usepackage{threeparttable}
\usepackage[capitalize]{cleveref}
\crefname{section}{Sec.}{Secs.}
\crefname{table}{Tab.}{Tabs.}
\crefname{assumption}{Assump.}{Assumps.}
\crefname{problem}{Prob.}{Probs.}
\crefname{definition}{Defn.}{Defns.}

\Crefname{section}{Section}{Sections}
\Crefname{table}{Table}{Tables}
\Crefname{assumption}{Assumption}{Assumptions}

\newtheorem{problem}{Problem}

\title{Supervised Algorithmic Fairness in Distribution Shifts: A Survey
}

\author{
Minglai Shao$^{1}$\thanks{Equal contribution.}\thanks{Yujie Lin and Minglai Shao are corresponding authors.}
\and
Dong Li$^{2}$\footnotemark[1]\and
Chen Zhao$^{3}$\footnotemark[1]\and
Xintao Wu$^4$\and
Yujie Lin$^1$\footnotemark[2]\and
Qin Tian$^2$
\affiliations
$^1$School of New Media and Communication, Tianjin University, China\\
$^2$College of Intelligence and Computing, Tianjin University, China\\
$^3$Department of Computer Science, Baylor University, USA\\ 
$^4$Electrical Engineering and Computer Science Department, University of Arkansas,  USA\\
\emails
\{linyujie\_22, ld2022244154, tianqin123, shaoml\}@tju.edu.cn,\\
chen\_zhao@baylor.edu,
xintaowu@uark.edu
}
\begin{document}

\maketitle

\begin{abstract}
    Supervised fairness-aware machine learning under distribution shifts is an emerging field that addresses the challenge of maintaining equitable and unbiased predictions when faced with changes in data distributions from source to target domains.
In real-world applications, machine learning models are often trained on a specific dataset but deployed in environments where the data distribution may shift over time due to various factors. 
This shift can lead to unfair predictions, disproportionately affecting certain groups characterized by sensitive attributes, such as race and gender. 
In this survey, we provide a summary of various types of distribution shifts and comprehensively investigate existing methods based on these shifts, highlighting six commonly used approaches in the literature.
Additionally, this survey lists publicly available datasets and evaluation metrics for empirical studies.
We further explore the interconnection with related research fields, discuss the significant challenges, and identify potential directions for future studies.

\end{abstract}

\section{Introduction}
    \label{sec:intro}
    Fairness in machine learning has emerged as a crucial consideration in real-world applications, recognizing the societal impact of algorithmic decision-making. 
The significance of fairness is particularly evident in scenarios, such as hiring processes, loan approvals, and criminal justice systems, where biased algorithms can perpetuate and exacerbate existing inequalities.
Fairness in machine learning refers to the equitable treatment of individuals, irrespective of their sensitive characteristics, such as race and gender. 
It emphasizes the need to mitigate algorithmic discrimination and promote equal opportunities in model outcomes. 
Nevertheless, achieving fairness is not devoid of challenges, especially in the presence of distribution shifts.
These shifts can pose significant hurdles as models trained on source distributions may not generalize well to target data distributions, potentially exacerbating biases and undermining the intended fairness objectives. 
Addressing these challenges is essential for advancing the responsible and ethical deployment of machine learning systems in the real world.

% various distribution shifts, and fairness
There are two main lines of distribution shifts: general and fairness-specific distribution shifts. 
The former focuses on shifts involving the input features and labels.
Covariate shift \cite{shimodaira2000improving} refers to variations due to differences between the set of marginal distributions over instances. 
Label shift \cite{wang2003mining}, just as its name implies, indicates the changes in the distribution of the class variable. 
Concept shift \cite{conceptshift} refers to ``functional relation change" \cite{yamazaki2007asymptotic} due to the change amongst the instance-conditional distributions. 
On the other hand, approaches addressing fairness-specific shifts consider sensitive attributes, recognizing their significant correlation with fair training.
Demographic shift \cite{giguere2022fairness} refers to certain sensitive population subgroups becoming more or less probable during inference. 
Dependence shift \cite{roh2023improving} captures the correlation change between labels and sensitive attributes. 
Within these distribution shifts, the fairness of the trained model is directly impacted and may deteriorate when adapted to target domains.

To enhance the performance of fairness generalization under distribution shifts, in this survey, we thoroughly examine existing supervised fairness-aware machine learning methods and highlight six commonly used approaches in the literature. 
Methods falling under feature disentanglement \cite{zhao2023towards} and data augmentation \cite{pham2023fairness} focus on capturing invariance across various domains. 
Examining causal graphs and paths \cite{yao2023understanding} aids in comprehending how the model's predictions for different sensitive groups are generated, thereby identifying and addressing potential unfair factors.
Adjusting weights of instances or features is a pre-processing method \cite{roh2023improving} that enhances model performance by rectifying unfairness in the target domain. The objective of robust optimization \cite{du2021fair} is to minimize the worst-case loss over various subsets of the training set or other well-defined perturbation sets around the data. Regularization-based methods~\cite{an2022transferring} can decrease the correlation between representation and sensitive attributes. It can also enable the transfer of fairness across different domains~\cite{schumann2019transfer}.

% These methods have been made from a diverse set of directions, including feature disentanglement \textcolor{red}{[citations]}, data augmentation \textcolor{red}{[citations]}, causal inference \textcolor{red}{[citations]}, reweighting \textcolor{red}{[citations]}, and regularization-based approaches \textcolor{red}{[citations]}. 

% In this survey, we comprehensively investigate the existing supervised fairness-aware machine learning methods within the context of various distribution shifts. 
Our main contributions of this survey are summarized:
\begin{itemize}[leftmargin=*]
    \item We summarize a list of different types of distribution shifts and illustrate the effectiveness of generalizing a fairness-aware classifier from source to target domains in the context of each distribution shift.
    \item We categorize existing methods based on different distribution shifts and highlight six main approaches commonly used for handling such shifts.
    \item We compile a list of the publicly available datasets and survey the literature to identify the most commonly used evaluation metric for quantifying fairness.
    \item We point out the significant challenges and explore several future directions for study fairness under distribution shifts.
\end{itemize}
% The remainder of this survey is structured as follows.
% In \cref{sec:background}, we establish the foundation of an overview of fairness-aware machine learning under distribution shifts and compile a list summarizing various types of shifts. 
% We categorize existing approaches in \cref{sec:methods} and discuss the technical specifics of each category. 
% Datasets for empirical studies and evaluation metrics used in the literature are detailed in \cref{sec:dataset}.
% We discuss the challenges of current approaches and future directions in \cref{sec:discussions}.
% The paper is concluded in \cref{sec:conclusion}.

\section{Background}
    \label{sec:background}
    % \subsection{Notations} 
Let $\mathcal{X}\subseteq\mathbb{R}^p$ denotes a feature space. $\mathcal{Z}\subset\mathbb{Z}$ is a sensitive space. $\mathcal{Y}\subset\mathbb{Z}$ is defined as an output or a label space. 
In this survey, we narrow the scope and only concentrate on classification tasks.
A domain is defined as a joint distribution $\mathbb{P}_{XZY}:=\mathbb{P}(X,Z,Y)$ on $\mathcal{X}\times\mathcal{Z}\times\mathcal{Y}$. 
A dataset sampled \textit{i.i.d.} from a domain $\mathbb{P}_{XZY}$ is represented as $\mathcal{D}=\{(\mathbf{x}_i,z_i,y_i)\}_{i=1}^{|\mathcal{D}|}$, where $\mathbf{x},z,y$ are the realizations of random variables $X,Z,Y$ in the corresponding spaces.
A classifier parameterized by $\boldsymbol{\theta}\in\Theta$ in a the space $\mathcal{F}$ is denoted as $f_{\boldsymbol{\theta}}:\mathcal{X}$
% \footnote{The classifier $f$ can have a varying input space, which depends on the type of fairness being considered. In group fairness scenarios, the input space is denoted as $\mathcal{X}$, while in the context of learning fair representations, it extends to $\mathcal{X}\times\mathcal{Z}$.}
$\rightarrow\mathcal{Y}$.
We denote $\mathcal{E}_{src}$ and $\mathcal{E}_{tgt}$ as sets of domain labels for source and target domains, respectively.
% Throughout this survey, lowercase boldface letters represent vectors, lowercase italic letters denote scalars, uppercase calligraphic letters denote sets, and variables are indicated by capital italic letters.

\subsection{Supervised Algorithmic Fairness} 

Algorithmic fairness in supervised machine learning refers to the concern of ensuring that the predictions and decision-making made by a model do not perpetuate existing biases and mitigate discriminatory practices.
% in the data used for training. 
% Mathematically, this can be formulated as minimizing the disparity between the predicted outcomes for different demographic groups. 
% Let $X$ be the feature variable,  $Z$ be the sensitive attribute representing a protected group (\textit{e.g.,} gender and race), and $Y$ be the label.
A fair algorithm aims to seek a classifier trained using data sampled from a source domain that can be generalized to a target domain, ensuring robust predictive performance in model utility (\textit{e.g.,} accuracy) while maintaining fair dependence between outcomes $f_{\boldsymbol{\theta}}(X)$ and the sensitive attribute $Z$.
% \subsection{Fairness Notions}
A crucial aspect of generalizing model fairness 
% from source domains to a target domain 
is to regulate the $(f_{\boldsymbol{\theta}}(X),Y)$-correlation 
% between the model outcome and sensitive attributes, 
assessed by fairness notions, 
% during the training process. Fairness notions 
which can be broadly categorized into three types: group fairness (\textbf{G}), individual fairness (\textbf{I}), and counterfactual fairness (\textbf{CF}). 

\textbf{Group fairness}\footnote{For simplicity, we present notions of group fairness with binary classes and a single binary sensitive attribute.} 
aims to ensure equitable outcomes for different demographic groups characterized by sensitive attributes. This is often expressed through the lens of demographic parity (DP) and equalized odds (EO) \cite{hardt2016equality}, where the conditional probability of a positive outcome for each class is equal across different sensitive subgroups. 
\begin{equation}
\small
    \begin{aligned}
        \text{DP:} \quad &\mathbb{P}(f_{\boldsymbol{\theta}}(X)=1|Z=1) = \mathbb{P}(f_{\boldsymbol{\theta}}(X)=1|Z=0) \\
        \text{EO:} \quad &\mathbb{P}(f_{\boldsymbol{\theta}}(X)=1|Z=1, Y=y) \\
        &= \mathbb{P}(f_{\boldsymbol{\theta}}(X)=1|Z=0, Y=y), \forall y
    \end{aligned} 
    \label{eq:group fairness}
\end{equation}
Although numerous works showcase the effectiveness of generalizing fairness to target domains using group fairness, they fall short in capturing the goal of treating individual people in a fair manner \cite{dwork2012fairness}. This limitation has given rise to individual fairness.
% where $\hat{Y}=f_{\boldsymbol{\theta}}(X)$ represents predicted outcomes generated by $f_{\boldsymbol{\theta}}$.
% Although numerous works showcase the effectiveness of generalizing fairness to target domains using DP and EO, implementing such fairness notions has been observed to compromise the predictive accuracy of the classifier. To this end, given a loss function $\ell:\mathcal{Y}\times\mathcal{Y}\rightarrow\mathbb{R}$, the mini-max fairness \textcolor{red}{[citation]} is designed with the aim of achieving good performance across different sensitive groups. 
% \begin{equation}
% \small
%     \begin{aligned}
%         f_{\boldsymbol{\theta}}^*\in\arg\min_{f_{\boldsymbol{\theta}}\in\mathcal{F}}\max_{z\in\mathcal{Z}} \mathbb{E}_{\mathbb{P}_{X,Y|Z=z}}\ell(f_{\boldsymbol{\theta}}(X),Y)
%     \end{aligned} 
% \end{equation}

\textbf{Individual fairness} \cite{dwork2012fairness} focuses on treating similar individuals similarly. In contrast to group fairness, which aims to achieve fairness at the statistical level
% by ensuring that overall demographic proportions are reflected in outcomes
, individual fairness defines a metric with respect to the particular context and requires that instances that are similar according to such metric receive similar outcomes.
\begin{equation}
\small
    \begin{aligned}
        d[f_{\boldsymbol{\theta}}(\mathbf{x}_i),f_{\boldsymbol{\theta}}(\mathbf{x}_j)] \leq \delta\cdot D[\mathbf{x}_i,\mathbf{x}_j], \quad i\neq j
    \end{aligned}
\end{equation}
where $\delta>0$ is a constant, $d:\mathbb{R}\times\mathbb{R}\rightarrow\mathbb{R}$ and $D:\mathcal{X}\times\mathcal{X}\rightarrow\mathbb{R}$ denote suitable distance metrics in output and feature spaces, respectively.

% is a distance measurement, and $D:\mathcal{X}\times\mathcal{X}\rightarrow\mathbb{R}$ is a similarity metric for a supervised task.

% Mini-max fairness can be expressed as the minimization of the maximum disadvantage or deprivation experienced by any individual or group in a given distribution.

\textbf{Counterfactual fairness} \cite{kusner2017counterfactual} is often expressed through the use of causal inference and potential outcomes. It involves evaluating whether a model's predictions would remain fair if certain characteristics of an individual were different while keeping other factors constant. This approach aims to uncover and address unfairness in predictive models by exploring hypothetical situations to ensure equitable outcomes.
\begin{equation}
\small
    \begin{aligned}
        &\mathbb{P}(f_{{\boldsymbol{\theta}},Z\leftarrow z}(U)=y|X=\mathbf{x},Z=z) \\
        = \: &\mathbb{P}(f_{{\boldsymbol{\theta}},Z\leftarrow z'}(U)=y|X=\mathbf{x},Z=z), \forall y \:\text{and}\: z\neq z'
    \end{aligned}
\end{equation}
where $U$ denotes latent variables in a given causal model and $f_{{\boldsymbol{\theta}},Z\leftarrow z}(U)$ represents the solution for outcomes for a given $U=u$ where the equation for $Z$ are placed with $Z=z$.

In traditional fair machine learning, while many approaches \cite{corbett2018measure} have proven successful, their methods typically assume that source and target data originate from identical distributions. However, it is more realistic to expect methods to operate in non-stationary environments, where the distributions for source and target domains undergo shifts.

\subsection{Learning Fairness under Distribution Shifts}
In the context of distribution shift, given access to one or more distinct source domains $\{\mathbb{P}^s_{XZY}\}_{s=1}^{S}$, where $S=|\mathcal{E}_{src}|$ represents the number of source domains, the objective of learning fairness under distribution shift is to train a fair classifier $f_{\boldsymbol{\theta}}$ using source data. 
This classifier can generalize well to a distinct target domain $\mathbb{P}^{t}_{XZY}$ with respect to predicted model utility and fairness, where $t\in\mathcal{E}_{tgt}$ and $\mathbb{P}^{t}_{XZY}\neq\mathbb{P}^s_{XZY},\forall s\in\mathcal{E}_{src}$. 
% The desired outcome includes achieving good predictive performance in model utility while maintaining fair dependence between outcomes and sensitive attributes on target domains.
% the prediction error as well as the dependence between outcomes and sensitive attributes on \textcolor{red}{an unseen} target domain $\mathbb{P}^{t}_{XZY}$, is minimized, where $t\in\mathcal{E}\backslash\mathcal{E}_{src}$ and $\mathbb{P}^{t}_{XZY}\neq\mathbb{P}^s_{XZY},\forall s\in\mathcal{E}_{src}$. 

\begin{figure*}[t]
    \centering
    \includegraphics[width=0.95\linewidth]{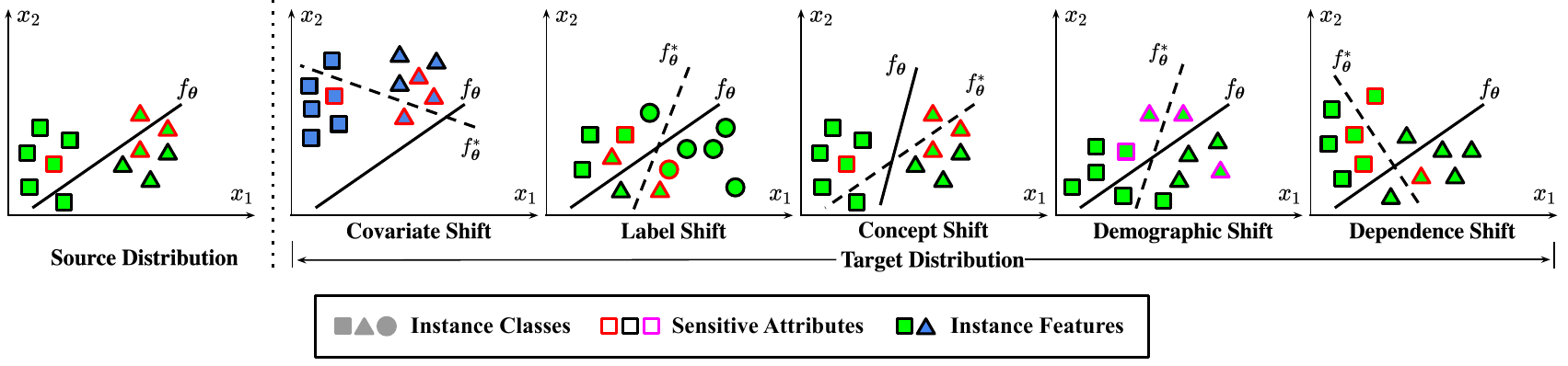}
    \caption{
    An illustration of fairness-aware machine learning under various distribution shifts. 
    We consider $S=1$ and $\mathbf{x}=[x_1,x_2]^T$ as a simple example of a two-dimensional feature vector. 
    \textbf{(Left)} A fair classifier $f_{\boldsymbol{\theta}}$ is learned using data sampled from a source domain. 
    \textbf{(Right)} The learned $f_{\boldsymbol{\theta}}$ is applied to data sampled from various types of shifted target domains, resulting in misclassification and unfairness. $f_{\boldsymbol{\theta}}^*$ represents the true classifier in the target domain.
    }
    \label{fig:shifts}
\end{figure*}

\begin{problem}[Learning Fairness under Distribution Shifts]
\label{prob:prob1}
    Let $\mathcal{D}_{src}=\{\mathcal{D}^s\}_{s=1}^{S}$ be a finite set of source data and assume that for each $s\in\mathcal{E}_{src}$, we have access to its corresponding data $\mathcal{D}^s=\{(\mathbf{x}^s_i,z^s_i,y^s_i)\}_{i=1}^{|\mathcal{D}^s|}$ sampled \textit{i.i.d} from its corresponding domain $\mathbb{P}^s_{XZY}$. Given a loss function $\ell:\mathcal{Y}\times\mathcal{Y}\rightarrow\mathbb{R}$, the goal is to learn a fair classifier $f_{\boldsymbol{\theta}}:\mathcal{X}\rightarrow\mathcal{Y}$ using $\mathcal{D}_{src}$ that simultaneously minimizes the prediction error and mitigates the dependence between predictive outcomes and sensitive attributes.
    The learned $f_{\boldsymbol{\theta}}$ can be further applied to a target dataset $\mathcal{D}_{tgt}$, sampled form a distinct target domain $\mathbb{P}^t_{XZY}$ that differs from all source domains where $\mathbb{P}^{t}_{XZY}\neq\mathbb{P}^s_{XZY},\forall s\in\mathcal{E}_{src}$.
    \begin{equation}
    \small
        \begin{aligned}
            \min_{\boldsymbol{\theta}\in\Theta}\: \mathbb{E}_{\mathbb{P}^{s}_{XZY},\forall s} \ell(f_{\boldsymbol{\theta}}(X^s),Y^s),\: \text{s.t.}\: \mathbb{E}_{\mathbb{P}^{s}_{XZY},\forall s} g(f_{\boldsymbol{\theta}}(X^s),Z^s)\leq \epsilon
        \end{aligned}
    \end{equation}
    where $g:\mathcal{Y}\times\mathcal{Z}\rightarrow\mathbb{R}$ denotes a fairness notion, describing the dependence between model outcomes and sensitive attributes. $\epsilon\geq 0$ is a threshold, which specifies an upper bound on the fair dependence and trades off model utility and fairness.
    % Notice that, in group fairness settings, the classifier is defined as $f:\mathcal{X}\rightarrow\mathcal{Y}$, whereas in the context of learning fair representations, it is denoted as $f:\mathcal{X}\times\mathcal{Z}\rightarrow\mathcal{Y}$. 
\end{problem}
% \textcolor{red}{[notes]}

As stated in \cref{prob:prob1}, a major challenge is to train a $f_{\boldsymbol{\theta}}$ that can be well generalized to a target domain from source domains under certain distribution shifts. 
In the following subsection, we outline various types of shifts and offer a brief analysis regarding fairness generalization for each shift.

    \begin{table}[t]
    \centering
    \setlength\tabcolsep{2pt}
    \scriptsize
    
    \begin{tabular}{lll}
        \toprule
         \textbf{Type of Shifts} & \textbf{Notations}, $\forall s\in\mathcal{E}_{src}$ & \textbf{References}\\
        \midrule
          Covariate Shift  & $\mathbb{P}_{X}^s\neq\mathbb{P}_{X}^t$ & \cite{shimodaira2000improving}\\
          \rowcolor{gray!15} Label Shift  & $\mathbb{P}_{Y}^s\neq\mathbb{P}_{Y}^t$ & \cite{wang2003mining}\\
          Concept Shift  & $\mathbb{P}_{Y|X}^s\neq\mathbb{P}_{Y|X}^t$ & \cite{conceptshift}\\
          \rowcolor{gray!15} Demographic Shift   & $\mathbb{P}_{Z}^s\neq\mathbb{P}_{Z}^t$ & \cite{giguere2022fairness}\\
           \multirow{2.4}{*}{Dependence Shift } & $\mathbb{P}_{Y|Z}^s\neq\mathbb{P}_{Y|Z}^t$ and $\mathbb{P}_{Z}^s=\mathbb{P}_{Z}^t$; & \multirow{2.4}{*}{\cite{roh2023improving}}\\
            & or $\mathbb{P}_{Z|Y}^s\neq\mathbb{P}_{Z|Y}^t$ and $\mathbb{P}_{Y}^s=\mathbb{P}_{Y}^t$ & \\
           \rowcolor{gray!15} Hybrid Shift & \multicolumn{2}{l}{Any combination of the shifts above, see \cref{tab:table_TASKS}} \\
        \bottomrule
    \end{tabular}
    \label{tab:shifts}
    \caption{Different Types of Distribution Shifts.}
\end{table}

\subsection{Different Types of Distribution Shifts}
An overview of different types of distribution shifts is summarized in \cref{tab:shifts} and \cref{fig:shifts} with illustrative examples.

\textbf{Covariate shift} \cite{shimodaira2000improving} 
refers to changes in the marginal distribution of the input variable $X$, formally $\mathbb{P}^s_X\neq\mathbb{P}^t_X,\forall s\in\mathcal{E}_{src}$. 
When the covariate shift occurs, the model's assumptions about the relationships between $X$ and $Y$ may no longer hold, leading to biased predictions. 
This bias disproportionately affects sensitive subgroups, especially if the training data does not adequately represent their protected characteristics. 
Consequently, the model may exhibit unfair behavior by making predictions that systematically disadvantage or advantage specific to these subgroups, reinforcing or exacerbating demographic disparities. 

\textbf{Label shift} \cite{wang2003mining}
, also known as prior probability shift \cite{saerens2002adjusting} or semantic shift \cite{newman2015semantic}, refers to changes in the distribution of the class variable $Y$, denoted as $\mathbb{P}^s_Y\neq\mathbb{P}^t_Y,\forall s\in\mathcal{E}_{src}$. 
The challenge of fairness learning under label shift primarily involves mitigating predictive bias in studies related to outlier or out-of-distribution detection. 
In such cases, samples with unknown classes (\textit{i.e.,} circles in \cref{fig:shifts}) are present only in the target domain. 
As a consequence of these samples being unknown to the learned fair classifier $f_{\boldsymbol{\theta}}$, it struggles to preserve the fair correlation between outcomes and sensitive variables while simultaneously maintaining prediction accuracy in distinguishing between known and unknown samples.

\textbf{Concept shift} \cite{conceptshift} is defined as changes in among the instance-conditional distributions, expressed as $\mathbb{P}^s_{Y|X}\neq\mathbb{P}^t_{Y|X},\forall s\in\mathcal{E}_{src}$.
It happens when the distribution of $Y$ given $X$ changes, which presents the hardest challenge among the different types of $(X,Y)$-related shifts that have been tackled so far.
In the presence of a structural causality graph between $X$, $Z$, and $Y$ where $Z\rightarrow(X,Y)$ and $X\rightarrow Y$, a concept shift from source to target domains leads to a change of the conditional distribution $\mathbb{P}_{Y|Z}$ through $X$.
As a result, the generalization of the fair classifier fails.

\begin{table*}[t]
    \scriptsize
    \centering
    \setlength\tabcolsep{3pt}
    
    \begin{threeparttable}
    \rowcolors{5}{white}{gray!15}
    \begin{tabular}{llcccccccccccc}
        \toprule
                 \multirow{2.4}{*}{\textbf{}} & \multirow{2.4}{*}{\textbf{Reference}}  & \multicolumn{5}{c}{\textbf{Distribution Shifts}\tnote{2}} & \multicolumn{3}{c}{\textbf{Spaces Change}, \: $s\rightarrow t,\forall s\in\mathcal{E}_{src}$} & 
                \multirow{2.4}{*}{$S>1$}
                &\multirow{2.4}{*}{\makecell[c]{\textbf{Accessibility}\\\textbf{of} $\mathcal{D}_{tgt}$}} & \multirow{2.4}{*}{\makecell[c]{\textbf{Fairness}\\\textbf{Type}\tnote{3}}}
                 & \multirow{2.4}{*}{\makecell[c]{\textbf{Method}}}\\
                \cmidrule(lr){3-7} \cmidrule(lr){8-10}
                  & & \textbf{Cov.} & \textbf{Lab.} & \textbf{Con.} & \textbf{Dem.} & \textbf{Dep.} & $\mathcal{X}^s\neq\mathcal{X}^t$ & $\mathcal{Y}^s\neq\mathcal{Y}^t$ & $\mathcal{Z}^s\neq\mathcal{Z}^t$  \\
                \midrule
                % \midrule
                % covariate
                % \cite{coston2019fair} &\textcolor{Cerulean}{\ding{51}}  &  &  &  &  & \textcolor{Cerulean}{\ding{51}}& & &  &\textcolor{Cerulean}{\ding{51}} &G & RW \\
                % \cite{mandal2020ensuring} &\textcolor{Cerulean}{\ding{51}}  &  &  &  &  & & & &  &  &G &\\
                & \cite{taskesen2020distributionally} &\textcolor{Cerulean}{\ding{51}}  & \textcolor{gray}{\ding{55}} & \textcolor{gray}{\ding{55}}  &  \textcolor{gray}{\ding{55}} &  \textcolor{gray}{\ding{55}} & \textcolor{gray}{\ding{55}} & \textcolor{gray}{\ding{55}} & \textcolor{gray}{\ding{55}} & \textcolor{gray}{\ding{55}} & \textcolor{gray}{\ding{55}} &G &RO\\
                \cellcolor{white}&\cite{rezaei2020fairness} &\textcolor{Cerulean}{\ding{51}}  & \textcolor{gray}{\ding{55}} & \textcolor{gray}{\ding{55}} & \textcolor{gray}{\ding{55}} & \textcolor{gray}{\ding{55}} & \textcolor{gray}{\ding{55}} & \textcolor{gray}{\ding{55}} & \textcolor{gray}{\ding{55}} & \textcolor{gray}{\ding{55}} & \textcolor{gray}{\ding{55}} &G & RO\\
                &\cite{creager2020causal} &\textcolor{Cerulean}{\ding{51}}  & \textcolor{gray}{\ding{55}} & \textcolor{gray}{\ding{55}} & \textcolor{gray}{\ding{55}} & \textcolor{gray}{\ding{55}} & \textcolor{gray}{\ding{55}} & \textcolor{gray}{\ding{55}} & \textcolor{gray}{\ding{55}} &\textcolor{Cerulean}{\ding{51}} & \textcolor{gray}{\ding{55}}  &G &CI\\
                  %\cite{shekhar2021fairod} &\textcolor{Cerulean}{\ding{51}}  & \textcolor{gray}{\ding{55}} & \textcolor{gray}{\ding{55}} & \textcolor{gray}{\ding{55}} & \textcolor{gray}{\ding{55}} & \textcolor{gray}{\ding{55}} & \textcolor{gray}{\ding{55}} & \textcolor{gray}{\ding{55}} & \textcolor{gray}{\ding{55}} & \textcolor{gray}{\ding{55}} &G &RG\\
                  
                  %\cite{abraham2021fairlof} &\textcolor{Cerulean}{\ding{51}}  & \textcolor{gray}{\ding{55}} & \textcolor{gray}{\ding{55}} & \textcolor{gray}{\ding{55}} & \textcolor{gray}{\ding{55}} & \textcolor{gray}{\ding{55}} &  \textcolor{gray}{\ding{55}} & \textcolor{gray}{\ding{55}} & \textcolor{gray}{\ding{55}} & \textcolor{gray}{\ding{55}} &G &OT\\
                  
                  %\cite{song2021deep} &\textcolor{Cerulean}{\ding{51}}  &  &  &  &  & & & &  &  &G &\\
                  \cellcolor{white}&\cite{rezaei2021robust} &\textcolor{Cerulean}{\ding{51}}  & \textcolor{gray}{\ding{55}} & \textcolor{gray}{\ding{55}} & \textcolor{gray}{\ding{55}} & \textcolor{gray}{\ding{55}} & \textcolor{gray}{\ding{55}} & \textcolor{gray}{\ding{55}} & \textcolor{gray}{\ding{55}} & \textcolor{gray}{\ding{55}} &\textcolor{Cerulean}{\ding{51}} &G & RW\\
                &\cite{du2021fair}
                  &\textcolor{Cerulean}{\ding{51}}  & \textcolor{gray}{\ding{55}} & \textcolor{gray}{\ding{55}} & \textcolor{gray}{\ding{55}} & \textcolor{gray}{\ding{55}} & \textcolor{gray}{\ding{55}} & \textcolor{gray}{\ding{55}} & \textcolor{gray}{\ding{55}} & \textcolor{gray}{\ding{55}} & \textcolor{Cerulean}{\ding{51}} &G & RO\\
                  \cellcolor{white}&\cite{zhao2021fairness} &\textcolor{Cerulean}{\ding{51}}  & \textcolor{gray}{\ding{55}} & \textcolor{gray}{\ding{55}} & \textcolor{gray}{\ding{55}} & \textcolor{gray}{\ding{55}} & \textcolor{gray}{\ding{55}} & \textcolor{gray}{\ding{55}} & \textcolor{gray}{\ding{55}} & \textcolor{gray}{\ding{55}} & \textcolor{gray}{\ding{55}} &G &RG\\
                  
                  %\cite{li2022fair} &\textcolor{Cerulean}{\ding{51}}  & \textcolor{gray}{\ding{55}} & \textcolor{gray}{\ding{55}} & \textcolor{gray}{\ding{55}} & \textcolor{gray}{\ding{55}} & \textcolor{gray}{\ding{55}} & \textcolor{gray}{\ding{55}} & \textcolor{gray}{\ding{55}} & \textcolor{gray}{\ding{55}} & \textcolor{gray}{\ding{55}} &G &OT\\
                  
                  &\cite{zhao2022adaptive} &\textcolor{Cerulean}{\ding{51}}  & \textcolor{gray}{\ding{55}} & \textcolor{gray}{\ding{55}} & \textcolor{gray}{\ding{55}} & \textcolor{gray}{\ding{55}} & \textcolor{Cerulean}{\ding{51}}& \textcolor{gray}{\ding{55}} & \textcolor{gray}{\ding{55}} & \textcolor{gray}{\ding{55}} & \textcolor{gray}{\ding{55}} &G & RW\\
                \cellcolor{white} &\cite{pham2023fairness} &\textcolor{Cerulean}{\ding{51}}  & \textcolor{gray}{\ding{55}} & \textcolor{gray}{\ding{55}} & \textcolor{gray}{\ding{55}} & \textcolor{gray}{\ding{55}} &\textcolor{Cerulean}{\ding{51}} & \textcolor{gray}{\ding{55}} & \textcolor{gray}{\ding{55}} &\textcolor{Cerulean}{\ding{51}} & \textcolor{gray}{\ding{55}} &G &DA, RG\\

                   &\cite{zhao2023towards} &\textcolor{Cerulean}{\ding{51}}  & \textcolor{gray}{\ding{55}} & \textcolor{gray}{\ding{55}} & \textcolor{gray}{\ding{55}} & \textcolor{gray}{\ding{55}} & \textcolor{Cerulean}{\ding{51}} & \textcolor{gray}{\ding{55}} & \textcolor{gray}{\ding{55}} & \textcolor{gray}{\ding{55}} & \textcolor{gray}{\ding{55}} &G & FD, RG\\
                  \cellcolor{white}\multirow{-10}{*}{\makecell[l]{\textbf{Covariate}\\ \textbf{Shift}}}&\cite{lin2023pursuing} &\textcolor{Cerulean}{\ding{51}}  & \textcolor{gray}{\ding{55}} & \textcolor{gray}{\ding{55}} & \textcolor{gray}{\ding{55}} & \textcolor{gray}{\ding{55}} & \textcolor{gray}{\ding{55}} & \textcolor{gray}{\ding{55}} & \textcolor{gray}{\ding{55}} & \textcolor{gray}{\ding{55}} & \textcolor{gray}{\ding{55}} &CF &FD, RG\\            

                  &\cite{li2024graph} &\textcolor{Cerulean}{\ding{51}}  & \textcolor{gray}{\ding{55}} & \textcolor{gray}{\ding{55}} & \textcolor{gray}{\ding{55}} & \textcolor{gray}{\ding{55}} & \textcolor{gray}{\ding{55}} & \textcolor{gray}{\ding{55}} & \textcolor{gray}{\ding{55}} & \textcolor{gray}{\ding{55}} & \textcolor{gray}{\ding{55}} &G & DA\\
                \midrule
                % Label
                  \cellcolor{white}\multirow{1}{*}{\makecell[c]{\textbf{Label Shift}}} & \cite{biswas2021ensuring} &   \textcolor{gray}{\ding{55}} &\textcolor{Cerulean}{\ding{51}} & \textcolor{gray}{\ding{55}} & \textcolor{gray}{\ding{55}} & \textcolor{gray}{\ding{55}} & \textcolor{gray}{\ding{55}} & \textcolor{gray}{\ding{55}} & \textcolor{gray}{\ding{55}} & \textcolor{gray}{\ding{55}} &\textcolor{Cerulean}{\ding{51}} &G & RO\\

                \midrule
                % Concpet 
                \cellcolor{white}\textbf{Concept} & \cite{iosifidis2019fairness} & \textcolor{gray}{\ding{55}}  &  \textcolor{gray}{\ding{55}} & \textcolor{Cerulean}{\ding{51}} & \textcolor{gray}{\ding{55}}  &  \textcolor{gray}{\ding{55}} & \textcolor{gray}{\ding{55}} & \textcolor{gray}{\ding{55}} & \textcolor{gray}{\ding{55}} & \textcolor{gray}{\ding{55}} & \textcolor{Cerulean}{\ding{51}} & G & RW\\
                \cellcolor{white}\textbf{Shift} &\cite{iosifidis2020online} &  \textcolor{gray}{\ding{55}} & \textcolor{gray}{\ding{55}}  & \textcolor{Cerulean}{\ding{51}} &  \textcolor{gray}{\ding{55}} & \textcolor{gray}{\ding{55}}  & \textcolor{gray}{\ding{55}} & \textcolor{gray}{\ding{55}} & \textcolor{gray}{\ding{55}} & \textcolor{gray}{\ding{55}} & \textcolor{Cerulean}{\ding{51}} & G & OT\\

                \midrule
                % Demographic 
                 \cellcolor{white}\textbf{Demographic} & \cite{schumann2019transfer} &   \textcolor{gray}{\ding{55}} & \textcolor{gray}{\ding{55}} &  \textcolor{gray}{\ding{55}}&\textcolor{Cerulean}{\ding{51}} & \textcolor{gray}{\ding{55}} & \textcolor{gray}{\ding{55}} & \textcolor{gray}{\ding{55}} &\textcolor{Cerulean}{\ding{51}} & \textcolor{gray}{\ding{55}} & \textcolor{Cerulean}{\ding{51}} &G &RG\\
                 \cellcolor{white}\textbf{Shift} &\cite{giguere2022fairness} &  \textcolor{gray}{\ding{55}} & \textcolor{gray}{\ding{55}} & \textcolor{gray}{\ding{55}} &\textcolor{Cerulean}{\ding{51}} & \textcolor{gray}{\ding{55}} & \textcolor{gray}{\ding{55}} & \textcolor{gray}{\ding{55}} &\textcolor{Cerulean}{\ding{51}} & \textcolor{gray}{\ding{55}}  & \textcolor{Cerulean}{\ding{51}} &G &OT\\

                 \midrule
                 % Dependence
                 \cellcolor{white} & \cite{creager2021environment} &  \textcolor{gray}{\ding{55}} & \textcolor{gray}{\ding{55}} & \textcolor{gray}{\ding{55}} & \textcolor{gray}{\ding{55}} &\textcolor{Cerulean}{\ding{51}} & \textcolor{gray}{\ding{55}} & \textcolor{gray}{\ding{55}} & \textcolor{gray}{\ding{55}} & \textcolor{Cerulean}{\ding{51}} & \textcolor{gray}{\ding{55}} &I & RO\\
                  \cellcolor{white}&\cite{oh2022learning} & \textcolor{gray}{\ding{55}} & \textcolor{gray}{\ding{55}} & \textcolor{gray}{\ding{55}} & \textcolor{gray}{\ding{55}} &\textcolor{Cerulean}{\ding{51}} & \textcolor{gray}{\ding{55}} & \textcolor{gray}{\ding{55}} & \textcolor{gray}{\ding{55}} & \textcolor{gray}{\ding{55}} & \textcolor{gray}{\ding{55}} &I & FD\\
                \cellcolor{white}\multirow{-3}{*}{\makecell[l]{\textbf{Dependence}\\ \textbf{Shift}}}&\cite{roh2023improving} & \textcolor{gray}{\ding{55}} & \textcolor{gray}{\ding{55}} & \textcolor{gray}{\ding{55}} & \textcolor{gray}{\ding{55}} &\textcolor{Cerulean}{\ding{51}} & \textcolor{gray}{\ding{55}} & \textcolor{gray}{\ding{55}} & \textcolor{gray}{\ding{55}} & \textcolor{gray}{\ding{55}} & \textcolor{Cerulean}{\ding{51}}  &G & RW\\

                \midrule
                % Hybrid
                 \cellcolor{white}&\cite{kallus2018residual} & \textcolor{Cerulean}{\ding{51}} & \textcolor{gray}{\ding{55}}  & \textcolor{Cerulean}{\ding{51}} & \textcolor{gray}{\ding{55}} & \textcolor{gray}{\ding{55}} & \textcolor{gray}{\ding{55}} & \textcolor{gray}{\ding{55}} & \textcolor{gray}{\ding{55}} & \textcolor{gray}{\ding{55}} & \textcolor{Cerulean}{\ding{51}} & G  &RW\\
                
                % \cite{singh2019fair} & \textcolor{Cerulean}{\ding{51}} & \textcolor{Cerulean}{\ding{51}} & \textcolor{Cerulean}{\ding{51}} & \textcolor{Cerulean}{\ding{51}} &   & & & &   & \textcolor{Cerulean}{\ding{51}} & G & CI \\
                % \cite{yoon2020joint} &   &   & \textcolor{Cerulean}{\ding{51}} & \textcolor{Cerulean}{\ding{51}} &   & & & \textcolor{Cerulean}{\ding{51}} &   & \textcolor{Cerulean}{\ding{51}} & G  & RO\\

                \cellcolor{white}&\cite{singh2021fairness} & \textcolor{Cerulean}{\ding{51}} & \textcolor{gray}{\ding{55}} & \textcolor{gray}{\ding{55}} & \textcolor{Cerulean}{\ding{51}} & \textcolor{gray}{\ding{55}}   & \textcolor{gray}{\ding{55}} & \textcolor{gray}{\ding{55}} & \textcolor{gray}{\ding{55}} & \textcolor{gray}{\ding{55}} & \textcolor{Cerulean}{\ding{51}} & G  & CI\\

                \cellcolor{white}&\cite{schrouff2022diagnosing} & \textcolor{Cerulean}{\ding{51}} & \textcolor{Cerulean}{\ding{51}} &  \textcolor{gray}{\ding{55}}& \textcolor{Cerulean}{\ding{51}} & \textcolor{gray}{\ding{55}}  & \textcolor{gray}{\ding{55}} & \textcolor{gray}{\ding{55}} & \textcolor{gray}{\ding{55}} & \textcolor{gray}{\ding{55}}   &\textcolor{Cerulean}{\ding{51}}  & G & CI\\
                
                 \cellcolor{white}&\cite{an2022transferring} &  \textcolor{Cerulean}{\ding{51}} & \textcolor{Cerulean}{\ding{51}} & \textcolor{gray}{\ding{55}}  & \textcolor{Cerulean}{\ding{51}} & \textcolor{Cerulean}{\ding{51}} & \textcolor{Cerulean}{\ding{51}} & \textcolor{gray}{\ding{55}} & \textcolor{gray}{\ding{55}} & \textcolor{gray}{\ding{55}} & \textcolor{Cerulean}{\ding{51}} & G  & RG, RW\\

                 \cellcolor{white}&\cite{chen2022fairness} &  \textcolor{Cerulean}{\ding{51}} & \textcolor{Cerulean}{\ding{51}} & \textcolor{gray}{\ding{55}}  & \textcolor{gray}{\ding{55}} & \textcolor{gray}{\ding{55}} & \textcolor{Cerulean}{\ding{51}} & \textcolor{gray}{\ding{55}} & \textcolor{gray}{\ding{55}} & \textcolor{gray}{\ding{55}} & \textcolor{Cerulean}{\ding{51}} & G  & RO\\

                 \cellcolor{white}&\cite{zhao2023fairness} & \textcolor{Cerulean}{\ding{51}} &  \textcolor{gray}{\ding{55}} & \textcolor{gray}{\ding{55}} & \textcolor{gray}{\ding{55}} & \textcolor{Cerulean}{\ding{51}} & \textcolor{Cerulean}{\ding{51}}  & \textcolor{gray}{\ding{55}} & \textcolor{gray}{\ding{55}} & \textcolor{Cerulean}{\ding{51}}  & \textcolor{gray}{\ding{55}} & G & FD, DA, RG\\
                 \cellcolor{white}\multirow{-7}{*}{\makecell[l]{\textbf{Hybrid}\\ \textbf{Shift}}}&\cite{han2023achieving} &\textcolor{Cerulean}{\ding{51}}  & \textcolor{Cerulean}{\ding{51}} & \textcolor{gray}{\ding{55}} & \textcolor{gray}{\ding{55}} & \textcolor{gray}{\ding{55}} & \textcolor{gray}{\ding{55}} & \textcolor{Cerulean}{\ding{51}} & \textcolor{gray}{\ding{55}} & \textcolor{gray}{\ding{55}} & \textcolor{gray}{\ding{55}} &CF &CI\\

        % \midrule
        \bottomrule
    \end{tabular}
    \begin{tablenotes}
        \item[2] The abbreviations stand for covariate (Cov.), label (Lab.), concept (Con.), demographic (Dem.), and dependence (Dep.) shifts.
        \item[3] The abbreviations stand for group (G), individual (I), and counterfactual (CF) fairness.
        \end{tablenotes}
    \end{threeparttable}
    \caption{An overview of existing approaches in fairness machine learning under various types of distribution shifts.}
    \label{tab:table_TASKS}
\end{table*}

\textbf{Demographic shift} \cite{giguere2022fairness} refers to changes in the marginal distribution of the sensitive variable $Z$ between source and target domains, denoted $\mathbb{P}^s_Z\neq\mathbb{P}^t_Z,\forall s\in\mathcal{E}_{src}$.
In particular, a demographic shift occurs when a specific sensitive subgroup of the population becomes more or less probable in the target domain.
As demonstrated in \cref{fig:shifts}, specifically in the context of the demographic shift, both source and target domains consist of samples with three sensitive attributes (indicated by distinct border colors in black, red, and magenta).
However, the number of samples indicated by magenta and red in the source and target domains is zero, respectively. 
Although $f_{\boldsymbol{\theta}}$ provides fair predictions in the source domain for the black and red groups, it might struggle to generalize fairness to the target domain in the presence of a demographic shift.

\textbf{Dependence shift} \cite{roh2023improving} can explicitly capture the alteration in the correlation between $Y$ and $Z$ of the data across the source and target domains.
In \cite{zhao2023towards}, the authors attribute the dependence shift to the changes in the conditional distribution involving $Y$ and $z$. Therefore, we formulate this shift with two similar alternative cases, $\mathbb{P}^s_{Y|Z}\neq\mathbb{P}^t_{Y|Z}$ and $\mathbb{P}^s_{Z}=\mathbb{P}^t_{Z}$, or $\mathbb{P}^s_{Z|Y}\neq\mathbb{P}^t_{Z|Y}$ and $\mathbb{P}^s_{Y}=\mathbb{P}^t_{Y}$.
Notice that 
% \citet{roh2023improving} 
\textit{Roh et al.,} \shortcite{roh2023improving} indicate handling dependence shift is straightforward when it is greater in the source domain than in the target domain. Otherwise, $f_{\boldsymbol{\theta}}$ may fail.

% \textbf{Hybrid Shift}

\textbf{Other types of shifts.} 
The aforementioned distribution shifts can be further categorized based on whether they undergo a corresponding change in space. The space change across domains requires a change in the corresponding distribution, but the opposite may not hold. For example, changes in the label space $\mathcal{Y}^s\neq\mathcal{Y}^t,\forall s$ and sensitive space $\mathcal{Z}^s\neq\mathcal{Z}^t,\forall s$ indicate the introduction of new labels and new sensitive attributes, which requires label shift and demographic shift, respectively. However, these shifts with invariant corresponding spaces refer to alterations in the proportions of instances generated from respective domains. A change in $\mathcal{X}$ denotes a shift in feature variation, such as transitioning from photo images to cartoons.

\section{Methods}
    \label{sec:methods}

We classify existing methods for addressing algorithmic fairness in distribution shifts into six categories: feature disentanglement (\textbf{FD}), data augmentation (\textbf{DA}), causal inference (\textbf{CI}), reweighting (\textbf{RW}), robust optimization (\textbf{RO}), and regularization-based approaches (\textbf{RG}). 
Any methods not falling within these categories are designated as others (\textbf{OT}).

% \subsection{Feature Disentanglement}
\textbf{Feature disentanglement} aims to learn latent representations of data features, enhancing their clarity and mutual independence within the model. The primary objective of this process is to enable the model to better comprehend the structure and variations present in the data. This endeavor involves disentangling the data representations into different latent spaces, facilitating a more nuanced understanding of the underlying data structure and dynamics. For the issue of fairness under distribution shifts, we obtain high-dimensional representations, denoted as $R = h(X)$, from the data, where $h:\mathcal{X}\rightarrow\mathcal{R}$ represents a high-dimensional mapping. 
This method aims to preserve only the semantic information unrelated to the sensitive attribute $Z$ for the final fair decision-making process. A commonly employed approach involves utilizing VAE-based methods for disentanglement~\cite{oh2022learning}, and some methodologies leverage structures resembling autoencoders~\cite{zhao2023fairness}. The advantage of feature disentanglement lies in its pronounced interpretability. However, directly evaluating the quality of the disentanglement poses challenges.

% \subsection{Data Augmentation}
\textbf{Data augmentation} aims to enhance the diversity of training datasets and improve model generalization performance by systematically applying controlled transformations to the training data. The fundamental idea behind this approach is to generate new samples that are similar but slightly different from the original data by applying various transformations. Additionally, one can also employ generative models (\textit{e.g.,} GANs) to train a generator~\cite{pham2023fairness} initially. This generator is designed to ensure that the generated representation $R$ satisfies the equality conditions: $\mathbb P_{R|Y}^i = \mathbb P_{R|Y}^j$ and $\mathbb P_{R|Y,Z}^i = \mathbb P_{R|Y,Z}^j$ for any source domains $\forall i,j \in\mathcal{E}_{src}$, $i\ne j$. 

Moreover, this generator can be utilized to generate synthetic data for the purpose of data augmentation. Data augmentation can also be combined with feature disentanglement. Initially, semantic factors, domain-specific factors, and sensitive factors are disentangled from the data. By randomly sampling a new set of domain-specific and sensitive factors from their prior distributions, along with the original semantic information, a decoder is employed to generate synthetic data~\cite{zhao2023fairness}. While data augmentation can enhance the diversity of the training set, enabling the model to generalize better to unseen target domains and thereby improving overall performance, its efficacy is highly contingent upon the chosen transformation strategies. 
% Different tasks and data types may necessitate distinct augmentation strategies. 

%In practical application, it is imperative to judiciously select and adjust data augmentation methods based on the specific problem and data characteristics to strike a balance between its advantages and disadvantages.

% \subsection{Causal Inference}
\textbf{Causal Inference.} 
Causal models have been widely applied in machine learning to address issues related to model fairness. Structural Causal Models (SCMs) provide a means of explaining machine learning model predictions. Analyzing causal graphs and paths helps understand how the model's predictions for different groups are formed, thereby identifying and addressing potential unfair factors. A typical fairness causal graph comprises three relational expressions: $Z\to X$, $Z\to Y$, and $Z, X\to Y$. In the context of various distribution shifts, this is equivalent to applying \textit{do}-operator to corresponding variables, where an intervention on $X$ (denoted $do(X)$) serves the purpose of resetting the distribution of $X$. One can apply the \textit{do}-operator separately to $X$, $Z$, and $Y$ to generate counterfactual samples, thereby calculating training sample influence on the model's unfairness~\cite{yao2023understanding}. Similarly, one can analyze the causal dependencies among $X$, $Z$, and $Y$ and their impact on the convergence speed of model training when transitioning to a new distribution~\cite{lin2023adaptation}. 
If only the $do(Z)$ operation is performed, the investigation of counterfactual fairness issues can be conducted. By setting $Z$ to a value different from its original state (\textit{e.g.,} transitioning from male to female), changes in $Y$ may be induced through the propagation of causal paths. It aims for the impact of $do(Z)$ on the predicted values to be minimized~\cite{han2023achieving}.

% \subsection{Reweighting}
% \textbf{Reweighting.} The reweighting approach \cite{calders2009building} is a commonly used pre-processing method in statistics and machine learning. It is employed to adjust the weights of samples or features to enhance model performance or correct biases in the dataset. For instance, deprived groups may receive higher scores compared to favored groups. These methods typically involve assigning weights to instances from the source domain to represent the overall distribution of the target domain, as class labels are only available for the source data. Due to their typically model-agnostic nature, after the pre-processing stage, this approach is applicable to any classifier. One of the related research employs a sample reweighting technique, where the impact of each instance is adjusted using the classical covariate shift instance weighing function to estimate accuracy metrics \cite{kallus2018residual}. This results in the spurious fairness policy being consistently and uniformly more stringent than necessary for the disadvantaged class. To improve fair training under dependence shift, a decoupling framework was introduced, wherein pre-processing is employed to modify the correlation between sensitive attributes and labels \cite{roh2023improving}. And then in-processing techniques are subsequently utilized to ensure equal weighting of samples within each $(y, z)$-class.

\textbf{Reweighting.} 
The reweighting approach is a commonly used pre-processing method in statistics and machine learning. It is employed to adjust the weights of samples or features to enhance model performance or correct biases in the dataset. These methods typically involve assigning weights to instances from the source domain to represent the overall distribution of the target domain. One related research employs a sample reweighting technique, where the impact of each instance is adjusted using the classical covariate shift instance weighing function to estimate accuracy metrics \cite{kallus2018residual}. To improve fair training under dependence shift, a decoupling framework was introduced, wherein pre-processing is employed to modify the correlation between sensitive attributes and labels \cite{roh2023improving}. And then in-processing techniques are subsequently utilized to ensure equal weighting of samples within each $(y, z)$-class.

% \subsection{Robust Optimization}
\textbf{Robust optimization.} 
A different strategy for extending the model's generalization beyond the training distribution is through robust optimization, wherein the objective is to minimize the worst-case loss over various subsets of the training set or other well-defined perturbation sets around the data. Most of these approaches operate within the framework of distributionally robust optimization (DRO). 
% Given a model family $\Theta$, loss function $\ell:\Theta\times\mathcal{X}\times\mathcal{Z}\times\mathcal{Y}\rightarrow\mathbb{R}$ and training sampled from distribution $\mathbb{P}^s_{XZY}$, 
DRO aims to minimize the worst-case training loss over any uncertainty set of distribution
% $\mathcal{Q}$
, which is proximate to the training distribution according to a specified metric.
% \begin{equation}
%     \textcolor{red}{\mathop{\text{min}}_{\theta\in\Theta}\mathop{\text{max}}_{Q\in\mathcal{Q}}\left\{\mathop{\text{sup}}_{Q\in\mathcal{Q}}\mathbb{E}_{(x,z,y)\sim Q}[\ell(\theta;(x,z,y))]\right\}.}
% \end{equation}

These methodologies are typically framed as a minimax problem, where the objective is to minimize the loss concerning the most adverse realization of perturbations within the uncertainty set
% $\mathcal{Q}$
. For instance, a double minimax iterative algorithm was introduced, wherein the set
% $\mathcal{Q}$ 
is defined by weighted perturbations of the empirical training distribution \cite{mandal2020ensuring}. \textit{Du and Wu} \shortcite{du2021fair} introduces a fairness constraint in both the minimization and maximization problems, with its uncertainty set defined as a Wasserstein ball centered around the uniform selection ratio. The uncertainty set also can be defined as the intersection of distributions within a Wasserstein ball centered at the empirical distribution \cite{taskesen2020distributionally}.

% \subsection{Regularization}
\textbf{Regularization-based approaches.} Fairness regularization can be broadly approached through two main strategies. The first involves introducing additional regularization terms to diminish the correlation between high-dimensional representations $R$ and sensitive attributes $Z$. For example, \textit{An et al.,} \shortcite{an2022transferring} adopts adversarial learning to hinder $R$ from encoding $Z$. Additionally, to ensure fairness and accuracy in the target domain under various shifts, the researchers utilize pseudo-labels generated by a teacher model as supervision for consistent training. This involves training the model to maintain consistent predictions under various transformations. Another approach involves employing regularization terms, such as Maximum Mean Discrepancy (MMD), to align distributions. In cases where the distribution of sensitive attributes $\mathbb P_Z$ differs between the source and target domains, with high-dimensional representations encoding sensitive attributes $\hat{Z}$ and domain labels $\hat{E}$, fairness transfer can be achieved by aligning the distributions between $\mathbb P_Z$ and $\mathbb P_{\hat{Z}}$, as well as $\mathbb P_E$ and $\mathbb P_{\hat{E}}$~\cite{schumann2019transfer}.

\begin{table*}[t]
    
    \centering
    \setlength\tabcolsep{3.5pt}
    \scriptsize
    % \rowcolors{5}{white}{gray!15}
    \begin{threeparttable}
        \begin{tabular}{l|llllllll}
            \toprule
            \multirow{2}{*}{\textbf{Type}} & \multirow{2}{*}{\textbf{Dataset}} & \multirow{2}{*}{$\#$ \textbf{Samples}} & \textbf{Feature} & \multirow{2}{*}{\textbf{Reference}} & \multirow{2}{*}{\textbf{Shifts}} & \textbf{Domain} & \textbf{Sensitive} &\multirow{2}{*}{\textbf{Class Label}} \\

             &  & & \textbf{Dimension} &  & \textcolor{red} &{\textbf{Characteristic}} & \textbf{Attribute} &\\
            \midrule

            % \texttt{Adult}
            \multirow{14}{*}{\textbf{Tabular}} & \multirow{4}{*}{\texttt{UCI Adult}} &\multirow{4}{*}{48,842}  & \multirow{4}{*}{14} &\cite{du2021fair} & Cov. &features   & gender  & income  \\
            &  &  & &\cellcolor{gray!15}\cite{han2023achieving}& \cellcolor{gray!15}Lab. & \cellcolor{gray!15}InD / OOD & \cellcolor{gray!15}gender  & \cellcolor{gray!15}InD / OOD \\
              &  &  & & \cite{iosifidis2020online} & Con. & time  &gender  & income \\ 
              &  &  & & \cellcolor{gray!15}\cite{yoon2020joint} & \cellcolor{gray!15}Dem. & \cellcolor{gray!15}gender \& race  &\cellcolor{gray!15}gender \& race  & \cellcolor{gray!15}income \\

            % \texttt{COMPAS}
            % \cmidrule{2-9}
             &    &  & &\cite{rezaei2021robust}& Cov. & features  &race  & recidivism \\

              &  && &\cellcolor{gray!15}\cite{han2023achieving}& \cellcolor{gray!15}Lab. & \cellcolor{gray!15}InD / OOD  & \cellcolor{gray!15}race  & \cellcolor{gray!15}InD / OOD \\
              
              & &  &  & \cite{biswas2021ensuring}&Lab.& recidivism  &race  & recidivism \\
              &\multirow{-4}{*}{\texttt{COMPAS}} & \multirow{-4}{*}{6,167}& \multirow{-4}{*}{9}& \cellcolor{gray!15}\cite{yoon2020joint}&\cellcolor{gray!15}Dem. & \cellcolor{gray!15}gender \& race  &\cellcolor{gray!15}gender \& race  & \cellcolor{gray!15}recidivism \\

            % \texttt{German}
            % \cmidrule{2-9}
             & \multirow{2}{*}{\texttt{German}} &\multirow{2}{*}{1,000}  & \multirow{2}{*}{20} & \cite{rezaei2021robust} & Cov. & features  & gender  & credit  \\
             
              &  &  & & \cellcolor{gray!15}\cite{yoon2020joint} & \cellcolor{gray!15}Dem. & \cellcolor{gray!15}gender \& age  &\cellcolor{gray!15}gender \& age  & \cellcolor{gray!15}credit \\

            % NYSF
            % \cmidrule{2-9}

            &   & 311,367  & 16 & \cite{iosifidis2020online} & Con. & years  & gender & sespect arrested  \\

            &  &  &  & \cellcolor{gray!15}\cite{kallus2018residual}& \cellcolor{gray!15}Cov.& \cellcolor{gray!15}features  & \cellcolor{gray!15}race  & \cellcolor{gray!15}weapon found \\
            
             % & \cellcolor{gray!15} &\cellcolor{gray!15}  & \cellcolor{gray!15} & \cellcolor{gray!15} &\cellcolor{gray!15}Cov. \& Dep. & \cellcolor{gray!15} city \& $(Y,Z)$-correlation  & \cellcolor{gray!15}race  & \cellcolor{gray!15}stop record  \\
             
              &\multirow{-3}{*}{\texttt{NYSF}}& \multirow{-2}{*}{685,724} & \multirow{-2}{*}{112} & \cite{zhao2023fairness} &Cov. \& Dep. & city \& $(Y,Z)$-correlation&race  & stop record \\

            %Credit Card~\cite{shekhar2021fairod} &24593
            %&1549  &  &  \\
   
            %Tweets~\cite{shekhar2021fairod} &3982  &10000  &   &  &  \\

            \midrule
            % CMNIST
            % \cmidrule{2-9}
            &\multirow{1}{*}{\texttt{cMNIST} }  &\multirow{1}{*}{70,000}  & \multirow{1}{*}{28$\times$28$\times$3} & \cellcolor{gray!15}\cite{creager2021environment} & \cellcolor{gray!15}Dep.  & \cellcolor{gray!15}$(Y,Z)$-correlation  & \cellcolor{gray!15}digit color  & \cellcolor{gray!15}digit group  \\

            % RCMNIST
            % \cmidrule{2-9}
             \cellcolor{white}&\cellcolor{white}\multirow{1}{*}{\texttt{rcMNIST}}  &\cellcolor{white}\multirow{1}{*}{10,000}  & \cellcolor{white}\multirow{1}{*}{28$\times$28$\times$3} & \cite{zhao2023towards}& Cov.  & rotated angle  & digit color  & digit group \\

            % \texttt{ccMNIST}
            % \cmidrule{2-9}
            &\multirow{1}{*}{\texttt{ccMNIST}} &\multirow{1}{*}{70,000}  & \multirow{1}{*}{28$\times$28$\times$3} & \cellcolor{gray!15}\multirow{1}{*}{\cite{zhao2023fairness}}& \cellcolor{gray!15}Cov. \& Dep.  & \cellcolor{gray!15}digit color \& $(Y,Z)$-correlation  & \cellcolor{gray!15}background color  & \cellcolor{gray!15}digit group \\
            
            % WaterBirds
            % \cmidrule{2-9}
             \cellcolor{white}&\cellcolor{white}\multirow{1}{*}{\texttt{Waterbirds}}  &\cellcolor{white}\multirow{1}{*}{4,795}  & \cellcolor{white}\multirow{1}{*}{not fixed} & \cite{creager2021environment} & Dep.  & $(Y,Z)$-correlation  & background  & bird’s breed \\

            % \texttt{FairFace}
            % \cmidrule{2-9}
            &\multirow{2}{*}{\texttt{FairFace}} &\multirow{2}{*}{108,501}  & \multirow{2}{*}{224$\times$224$\times$3} & \cellcolor{gray!15}\cite{an2022transferring} & \cellcolor{gray!15}Cov.  & \cellcolor{gray!15}data variation & \cellcolor{gray!15}race  & \cellcolor{gray!15}gender \\

            &&  & & \cite{zhao2023fairness} &  Cov. \& Dep.  & race \& $(Y,Z)$-correlation & gender  & age \\

            % \texttt{UTKFace}
            % \cmidrule{2-9}
           \rowcolor{gray!15} \cellcolor{white} \multirow{-7}{*}{\textbf{Image}}& \cellcolor{white}\multirow{1}{*}{\texttt{UTKFace}} &\cellcolor{white}\multirow{1}{*}{23,708}  & \cellcolor{white}\multirow{1}{*}{128$\times$128$\times$3} & \cite{an2022transferring} & Cov.  & data variation  & race  & gender \\

           \midrule

           \textbf{Text} & \texttt{BOLD} & 23,679 & N/A & \cite{dhamala2021bold} & Cov. & topic & gender & sentiment \\

           \midrule
           \multirow{3}{*}{\textbf{Graph}} & \texttt{Pokecs} & 134365 & 265 & \cellcolor{gray!15}\cite{li2024graph} & \cellcolor{gray!15}Cov. &\cellcolor{gray!15}province &\cellcolor{gray!15}region &\cellcolor{gray!15}working field
 \\

            & \texttt{Bail-Bs} & 12024 & 18 & \cite{li2024graph} & Cov. &  community & race & bail \\

            &\texttt{Credit-Cs} & 16009 & 13 &\cellcolor{gray!15}\cite{li2024graph} &\cellcolor{gray!15}Cov. &\cellcolor{gray!15}community &\cellcolor{gray!15}age &\cellcolor{gray!15}risk \\

            \bottomrule
        \end{tabular}%
        % \begin{tablenotes}
        % \item[5] Correlation between sensitive attribute and label.
        % \end{tablenotes}
    %}
    \end{threeparttable}
    \caption{Common datasets for fairness under distribution shifts.}
    \label{table:datasets}
\end{table*}

% \subsection{Other Methods}
\textbf{Other methods.} In addition to the methods mentioned above, there are other approaches to enhance the performance of fairness generalization under distribution shifts. Differing from reweighting methods employed during the preprocessing stage, \textit{decision boundary adjustment} \cite{iosifidis2020online} is introduced as a post-processing approach. This method ensures fairness under concept shift in an online learning setting. Cumulative discrimination is observed in a streaming environment, and the decision boundary is adjusted when it exceeds a threshold to meet fairness requirements. An universal training framework was developed based on the \textit{Seldonian algorithm} \cite{thomas2019preventing} to ensure fairness under demographic shifts. The training data was divided into two parts: one for training candidate fair classifiers using existing fairness algorithms and the other for calculating a high-confidence upper bound on the prevalence of unfair behavior. This allowed for the evaluation of candidate models to obtain models robust to demographic shifts.

\section{Datasets and Evaluation}
    \label{sec:dataset}
    
\subsection{Datasets}
% Many datasets have been utilized to assess fairness under distribution shifts. The same dataset, under different settings (domain partitions, sensitive attribute, and class label selections), can yield various distribution shifts, thus serving as a means to evaluate different scenarios. 
% We summarize the most commonly used publicly available datasets in \cref{table:datasets}.

% \subsubsection{Tabular Datasets}
\textbf{Tabular datasets.} 
\texttt{UCI Adult} \cite{kohavi1996scaling} includes nearly 48,842 adults and generates a binary income label by determining whether an individual's income exceeds $\$50$k. Gender is commonly chosen as the sensitive attribute.
% in fairness-aware algorithmic studies. 
\texttt{COMPAS} \cite{dressel2018accuracy} consists of 6,167 samples, and its objective is to predict an individual's recidivism based on criminal history. The dataset assigns race as the sensitive attribute. \texttt{UCI Adult} and \texttt{COMPAS} are also employed for fair anomaly detection, where the binary label indicates whether an instance is anomalous or not.
\texttt{German} \cite{asuncion2007uci} comprises a collection of 1,000 samples used for credit scoring prediction of whether an individual is a good or bad credit risk based on various financial features. The dataset includes a binary-sensitive attribute indicating the individual's gender along with other relevant features.
\texttt{New York Stop-and-Frisk (NYSF)} \cite{koh2021wilds} is a real dataset from the policing activities in New York City. It spans multiple years involving 685,724 records. Due to evident racial bias against African Americans, race is considered a sensitive attribute. Data in different cities serve as distinct domains, and various inspection records can be used as labels, such as stop record, weapon possession, or arrest status. \texttt{NYSF} is also utilized for online settings to address concept shift due to its temporal characteristics \cite{iosifidis2020online}. In this context, gender is the sensitive attribute.

\textbf{Image datasets.}
\texttt{Colored-MNIST (cMNIST)} \cite{arjovsky2019invariant} purposely establishes a correlation between digits and digit colors in the source data but intentionally anti-correlates them in the target one.
% The class is set to 0 when the digit belongs to $\{0,1,2,3,4\}$ and 1 when the digit belongs to $\{5,6,7,8,9\}$.
\texttt{Rotated-Colored-MNIST (rcMNIST)}~\cite{zhao2023towards} is extended from the \texttt{cMNIST}, 
% which consists of 10,000 digits 
where each digit is associated with different rotated angles representing domains.
% where environments are determined by angles $\{0^\circ, 15^\circ, 30^\circ, 45^\circ, 60^\circ, 75^\circ\}$. 
For fairness concerns, digit colors are the sensitive attribute correlated to labels. 
% To simplify, it considers only binary classification akin to \texttt{cMNIST}.
\texttt{Corlored-Corlored-MNIST (ccMNIST)} \cite{arjovsky2019invariant} is created by colorizing digits and the backgrounds of the \texttt{MNIST} dataset. \texttt{ccMNIST} contains three domains characterized by a different digit color with a different correlation between the class label (same as \texttt{cMNIST}) and sensitive attribute (background colors).
\texttt{Waterbirds} \cite{sagawa2019distributionally} is created by merging bird images with backgrounds. The label indicates the bird's breed, while the sensitive attribute corresponds to the background type. Similar to \texttt{cMNIST}, there exists a spurious correlation between the label and sensitive attribute, generating a dependence shift between the source and target domains. 
% Both datasets are employed in \cite{creager2021environment} to ensure fairness under dependence shift.
\texttt{FairFace} \cite{karkkainen2019fairface} contains 108,501 facial images across seven racial categories. A deployment for fairness in distribution shifts is to set the racial groups as domains, gender as the sensitive attribute, and age as class label \cite{zhao2023fairness}.
\texttt{UTKFace} \cite{zhang2017age} comprises 23,708 facial images. Each image is annotated with information regarding the subject's age, gender, and ethnicity. In this dataset, race is considered a sensitive attribute, while gender is the class label.

\textbf{Text datasets.}
\texttt{BOLD} \cite{dhamala2021bold} comprises 23,679 English text generation prompts designed to assess societal biases in open-ended language generation, specifically applied to sentiment analysis tasks. The prompts are categorized into five domains based on their topics: profession, gender, race, religion, and political ideology. Gender is considered a binary sensitive attribute within this context.

\textbf{Graph datasets.} \texttt{Pokecs}~\cite{dai2022learning} comprises 134,563 graphs extracted from a popular social network in Slovakia, organized according to the users' respective provinces. Different domains within the dataset are composed of users from two major regions within the corresponding provinces, with the region being treated as a sensitive attribute. \texttt{Bail-Bs} dataset, consisting of nodes representing defendants on bail, is derived from commonly used fair-related graphs obtained from \texttt{Bail}, with race being a sensitive attribute. \texttt{Credit-Cs} is derived from \texttt{Credit}, wherein nodes represent credit card users, following the same partitioning scheme as \texttt{Bail-Bs}. The task involves categorizing customers' credit risks as high or low based on age as the sensitive attribute.

% \textbf{Distribution Shifts in Datasets.}
% For tabular data, biased sampling can result in $\mathbb{P}^e_X \neq \mathbb{P}^t_X$, leading to covariate shift ~\cite{du2021fair,rezaei2021robust,kallus2018residual}. In anomaly detection, where the source data comprise only normal samples, and the target data introduce anomalous samples, label shift occurs \cite{han2023achieving}. Prior probability shifts between training and testing samples fundamentally represent a form of label shift \cite{biswas2021ensuring}. Demographic shift arises when source and target data use different sensitive attributes \cite{yoon2020joint}. Datasets such as \texttt{NYSF}, \texttt{ccMNIST}, and \texttt{FairFace} employ their attributes to delineate domains. This leads to differing correlations between $Z$ and $Y$ in the source and target data, giving rise to both covariate shift and dependence shift \cite{zhao2023fairness}. When \texttt{UTKFace} is used as source data and \texttt{FairFace} as target data, the variation in data also induces covariate shift \cite{an2022transferring}.

\subsection{Evaluation Metrics for Assessing Fairness}
%This section introduces the most commonly utilized algorithmic fairness metrics in machine learning classification tasks.
\begin{comment}
    We focus on three primary fairness issues: group fairness, individual fairness, and counterfactual fairness. To simplify the description, we assume here that both the sensitive attribute and the class label are binary labels. Subsequently, we elaborate on metrics specific to each of these three categories.
\end{comment}
\textbf{Metrics for Group Fairness.}
Three fundamental metrics are used for evaluating group fairness.
\textit{Difference of Demographic Parity} ($\Delta_{DP}$) \cite{dwork2012fairness} and \textit{Difference of Equalized Odds} ($\Delta_{EO}$) \cite{hardt2016equality} are similar to the notions presented in \cref{eq:group fairness}, where $\Delta_{DP}$ and $\Delta_{EO}$ take the absolute probability difference in DP and EO.
\textit{Difference of Equalized Opportunity} ($\Delta_{EOp}$)~\cite{hardt2016equality} is similar to $\Delta_{EO}$. The difference lies in that $\Delta_{EOp}$ only requires true positive rates (TPRs) across sensitive subgroups.
{
\small
\begin{align}
    \Delta_{EOp}= & | \mathbb{P}(f_{\boldsymbol{\theta}}(X)=1|Z=1,Y=1) \\
                  &-\mathbb{P}(f_{\boldsymbol{\theta}}(X)=1|Z=0,Y=1) | \nonumber
\end{align}
}
A value of zero for these metrics indicates fair predictions in a target domain.

\textbf{Metrics for Individual Fairness.}
% Individual Fairness emphasizes that similar individuals should receive similar treatment. 
Due to the need to measure the similarity between samples, there are relatively few relevant metrics, and \textit{consistency}~\cite{zemel2013learning} stands out as one of the more classical ones. Given an input $\mathbf x$ to its $k$-nearest neighbors, $kNN(\mathbf x)$, consistency is formulated:
{
\small
\begin{align}
    \text{consistency}=1-\frac{1}{Nk}\sum_{n}\left | f_{\boldsymbol{\theta}}(\mathbf{x}_n)-\sum_{j\in kNN(\mathbf x_n)} f_{\boldsymbol{\theta}}(\mathbf{x}_j) \right | 
\end{align}
}
where $N$ represents the total number of samples, and $\mathbf x_n$ denotes the $n$-th sample. A value close to $1$ represents fairness.

\textbf{Metrics for Counterfactual Fairness.}
As counterfactual fairness is a method defined by causal structures, one needs to use the $do$-operator to evaluate it. 
Assuming $A$ is the intervention target of the $do$-operator, $Y$ is influenced by this intervention. The post-intervention distribution of $Y$ can be expressed as $\mathbb P(y_a)=\mathbb P(Y=y|do(A=a))$. Currently, there are primarily two types of metrics.

The \textit{Total Causal Effect} (TCE)~\cite{pearl2009causality} of the value change of $Z$ from $z$ to $z'$ on $Y=y$ is given by
{
\small
\begin{align}
    \text{TCE}(z, z')= | \mathbb P(y_{z})-\mathbb P(y_{z'})|.
\end{align}
}
Given context $O=\mathbf{o}$, the 
\textit{Counterfactual Effect} (CE)~\cite{shpitser2008complete} of the value change of $Z$ from $z$ to $z'$ on $Y=y$ is given by
{
\small
\begin{align}
    \text{CE}(z,z'|\mathbf{o})=|\mathbb P(y_{z}|\mathbf{o})-\mathbb P(y_{z'}|\mathbf o)|.
\end{align}
}
Smaller TCE and CE indicate that the prediction results are more stable in the counterfactual generation of changing the sensitive attribute, implying greater fairness.
    
% \section{Discussions}
    % \label{sec:discussions}
    \section{Related Research Fields}

\textbf{Unsupervised algorithmic fairness under distribution shifts} focuses on addressing unfairness in machine learning models when data labels are not available, and the model needs to adapt to changes in the data distribution. 
As the currently few unsupervised fairness outlier detection methods, \textit{Song et al.}~\shortcite{song2021deep} leverage deep clustering to discover the intrinsic cluster structure and out-of-structure instances. 
Meanwhile, adversarial training erases the
sensitive pattern for instances of fairness adaptation. \textit{Coston et al.} \shortcite{coston2019fair} investigate unsupervised domain adaptation for covariate shift between a source and target distribution, highlighting the unavailability of information from the target domain. They employ reweighting to learn weights closely approximating covariate shift weights, defined solely by non-sensitive attributes available in both domains under statistical parity constraints on the source data. 

\textbf{Fairness-aware outlier detection.}
Unlike conventional classification tasks, outlier detection aims to identify samples where covariate shifts have occurred. Under the influence of sensitive attributes, samples from certain groups are more prone to being identified as outlier samples. \textit{Shekhar et al.}~\shortcite{shekhar2021fairod} decompose the fairness in outlier detection into two issues: DP and EOp. They design regularizations to address both problems. 
Based on the framework of LOF, FairLOF \cite{abraham2021fairlof} attempts to correct for such 
$k$NN neighborhood distance disparities across object groups defined over sensitive attributes. Similarly, based on LOF and adversarial networks, in contrast to FairLOF, AFLOF~\cite{li2022fair} learns the optimal representation of the original data while concealing the sensitive attribute in the data. %They introduce a dynamic weighting module that assigns lower weight values to data objects with higher local outlier factors to mitigate the influence of outliers on representation learning.

%\textbf{Fairness in reinforcement learning.}
%\textcolor{red}{more words...}

\section{Challenges and Future Directions} 
\textbf{Fairness under conditional shift.} In fact, it is deemed appropriate to align $\mathbb P_X$ only when $X$ is the causative factor of $Y$. 
\begin{comment}
    In such instances, $\mathbb P_{Y|X}$ remains stable despite variations in $\mathbb P_X$, as the two probabilities are not intricately linked.
\end{comment}
However, it is plausible that $Y$ serves as the cause of $X$, leading to an impact on $\mathbb P_{Y|X}$ when there are shifts in $\mathbb P_X$. The variation of $\mathbb P_{Y|X}$ in an unknown domain is referred to as \textit{conditional shift}~\cite{liu2021domain}. Consequently, certain domain alignment approaches~\cite{liu2021domain} have proposed aligning the class-conditional distribution $\mathbb P_{X|Y}$, under the assumption that $\mathbb P_Y$ remains unchanged. 
We believe fairness issues under conditional shift are a promising area for exploration.

% \textbf{Fairness under concept shift.} 
% %In the context of online learning, supervised learning encounters the challenge known as concept drift, wherein alterations in the underlying data distribution impact the learning model. This occurs due to the potential evolution of the relationship between input variables and class variables over time. 
% Existing solutions address concept shift by dynamically adjusting the learning model online. 
% However, the model's adaptability, leading to changes in the decision boundaries of classifiers, may compromise the fairness of the model. Some approaches address this issue by modifying input data~\cite{iosifidis2019fairness} or dynamically adjusting the model distribution online to align with fairness objectives~\cite{iosifidis2020online}. 
% However, in the conventional setting, it lacks a continuous data stream, making it challenging to discern the patterns of change in $\mathbb P_{Y|X}$. In cases where there is a significant concept shift between the source and target domains, addressing both fairness and accuracy in the target domain becomes formidable.

\textbf{Fairness in out-of-distribution detection.} 
%The objective of OOD detection is to empower the model to recognize and reject inputs originating from distributions beyond the training data, thus preventing the model from generating unreliable predictions in unknown scenarios. 
While existing work ~\cite{han2023achieving} ensures the absence of anomalous samples during the training process and guarantees fairness in anomaly detection during testing, out-of-distribution detection also necessitates addressing the classification of in-distribution samples. In other words, fairness concerns in out-of-distribution detection involve simultaneously considering multi-class fairness for in-distribution samples and binary-class fairness for out-of-distribution sample detection. Therefore, in the design of the model, it is imperative to balance fairness considerations for both aspects.

% \textbf{Multiple source domains.} The majority of the investigated works in this survey under consideration typically limit their scope to a single source domain. However, in real-world scenarios, data often exhibits a multifaceted origin, stemming from diverse sources. Learning across multiple source domains enables the model to acquire more robust representations, rendering it less sensitive to variations and noise inherent in a singular source domain.

% \textbf{Dataset specifically designed for fairness learning under distribution shifts.} While many datasets in the fields of algorithmic fairness or out-of-distribution scenarios have been utilized for fairness under distribution shifts, there is still a lack of a dataset explicitly designed for this task. Moreover, existing datasets lack a clear definition of the distribution shifts within them, creating a gap between algorithm design and experimental analysis. To address this issue more effectively, the design of a real-world dataset dedicated to this domain, capable of encompassing various types of distribution shifts, is urgently needed.

% multiple sensitive attribute，missing sensitive attribute，nultiple source domain

\section{Conclusion}
    \label{sec:conclusion}
    Ensuring the generalization of model fairness under distribution shifts across domains has emerged as a crucial research focus in recent years.
In this survey, we initially compile a list of diverse distribution shifts, offering a brief discussion on the reasons why a predicted model may fail to adapt to each setting.
Additionally, we offer a comprehensive review of existing methods addressing fairness generalization in various distribution shifts.
Finally, we identify several challenges and suggest potential avenues for future research.

%\section*{Contribution Statement}

\section*{Acknowledgments}
This work is supervised by Chen Zhao and Minglai Shao. This work is supported by the National Natural Science Foundation of China program (NSFC \#62272338).

\clearpage
\clearpage
\bibliographystyle{named}
\bibliography{reference}

\begin{thebibliography}{}

\bibitem[\protect\citeauthoryear{An \bgroup \em et al.\egroup }{2022}]{an2022transferring}
Bang An, Zora Che, Mucong Ding, and Furong Huang.
\newblock Transferring fairness under distribution shifts via fair consistency regularization.
\newblock {\em NeurIPS}, 35:32582--32597, 2022.

\bibitem[\protect\citeauthoryear{Arjovsky \bgroup \em et al.\egroup }{2019}]{arjovsky2019invariant}
Martin Arjovsky, L{\'e}on Bottou, Ishaan Gulrajani, and David Lopez-Paz.
\newblock Invariant risk minimization.
\newblock {\em arXiv:1907.02893}, 2019.

\bibitem[\protect\citeauthoryear{Asuncion and Newman}{2007}]{asuncion2007uci}
Arthur Asuncion and David Newman.
\newblock Uci machine learning repository, 2007.

\bibitem[\protect\citeauthoryear{Biswas and Mukherjee}{2021}]{biswas2021ensuring}
Arpita Biswas and Suvam Mukherjee.
\newblock Ensuring fairness under prior probability shifts.
\newblock In {\em AIES}, pages 414--424, 2021.

\bibitem[\protect\citeauthoryear{Chen \bgroup \em et al.\egroup }{2022}]{chen2022fairness}
Yatong Chen, Reilly Raab, Jialu Wang, and Yang Liu.
\newblock Fairness transferability subject to bounded distribution shift.
\newblock {\em NeurIPS}, 35:11266--11278, 2022.

\bibitem[\protect\citeauthoryear{Corbett-Davies and Goel}{2018}]{corbett2018measure}
Sam Corbett-Davies and Sharad Goel.
\newblock The measure and mismeasure of fairness: A critical review of fair machine learning.
\newblock {\em arXiv:1808.00023}, 2018.

\bibitem[\protect\citeauthoryear{Coston \bgroup \em et al.\egroup }{2019}]{coston2019fair}
Amanda Coston, Karthikeyan~Natesan Ramamurthy, Dennis Wei, Kush~R Varshney, Skyler Speakman, Zairah Mustahsan, and Supriyo Chakraborty.
\newblock Fair transfer learning with missing protected attributes.
\newblock In {\em AIES}, pages 91--98, 2019.

\bibitem[\protect\citeauthoryear{Creager \bgroup \em et al.\egroup }{2020}]{creager2020causal}
Elliot Creager, David Madras, Toniann Pitassi, and Richard Zemel.
\newblock Causal modeling for fairness in dynamical systems.
\newblock In {\em ICML}. PMLR, 2020.

\bibitem[\protect\citeauthoryear{Creager \bgroup \em et al.\egroup }{2021}]{creager2021environment}
Elliot Creager, J{\"o}rn-Henrik Jacobsen, and Richard Zemel.
\newblock Environment inference for invariant learning.
\newblock In {\em ICML}, pages 2189--2200. PMLR, 2021.

\bibitem[\protect\citeauthoryear{Dai and Wang}{2022}]{dai2022learning}
Enyan Dai and Suhang Wang.
\newblock Learning fair graph neural networks with limited and private sensitive attribute information.
\newblock {\em IEEE Transactions on Knowledge and Data Engineering}, 2022.

\bibitem[\protect\citeauthoryear{Deepak and Abraham}{2021}]{abraham2021fairlof}
Parakkal Deepak and Savitha~Sam Abraham.
\newblock Fairlof: fairness in outlier detection.
\newblock {\em Data Science and Engineering}, 2021.

\bibitem[\protect\citeauthoryear{Dhamala \bgroup \em et al.\egroup }{2021}]{dhamala2021bold}
Jwala Dhamala, Tony Sun, Varun Kumar, Satyapriya Krishna, Yada Pruksachatkun, Kai-Wei Chang, and Rahul Gupta.
\newblock Bold: Dataset and metrics for measuring biases in open-ended language generation.
\newblock In {\em FAccT}, pages 862--872, 2021.

\bibitem[\protect\citeauthoryear{Dressel and Farid}{2018}]{dressel2018accuracy}
Julia Dressel and Hany Farid.
\newblock The accuracy, fairness, and limits of predicting recidivism.
\newblock {\em Science advances}, 4(1):eaao5580, 2018.

\bibitem[\protect\citeauthoryear{Du and Wu}{2021}]{du2021fair}
Wei Du and Xintao Wu.
\newblock Fair and robust classification under sample selection bias.
\newblock In {\em CIKM}, pages 2999--3003, 2021.

\bibitem[\protect\citeauthoryear{Dwork \bgroup \em et al.\egroup }{2012}]{dwork2012fairness}
Cynthia Dwork, Moritz Hardt, Toniann Pitassi, Omer Reingold, and Richard Zemel.
\newblock Fairness through awareness.
\newblock In {\em Proceedings of the 3rd innovations in theoretical computer science conference}, 2012.

\bibitem[\protect\citeauthoryear{Giguere \bgroup \em et al.\egroup }{2022}]{giguere2022fairness}
Stephen Giguere, Blossom Metevier, Yuriy Brun, Bruno~Castro da~Silva, Philip~S Thomas, and Scott Niekum.
\newblock Fairness guarantees under demographic shift.
\newblock In {\em ICLR}, 2022.

\bibitem[\protect\citeauthoryear{Han \bgroup \em et al.\egroup }{2023}]{han2023achieving}
Xiao Han, Lu~Zhang, Yongkai Wu, and Shuhan Yuan.
\newblock Achieving counterfactual fairness for anomaly detection.
\newblock In {\em PAKDD}. Springer, 2023.

\bibitem[\protect\citeauthoryear{Hardt \bgroup \em et al.\egroup }{2016}]{hardt2016equality}
Moritz Hardt, Eric Price, and Nati Srebro.
\newblock Equality of opportunity in supervised learning.
\newblock {\em NeurIPS}, 29, 2016.

\bibitem[\protect\citeauthoryear{Iosifidis and Ntoutsi}{2020}]{iosifidis2020online}
Vasileios Iosifidis and Eirini Ntoutsi.
\newblock Fabboo-online fairness-aware learning under class imbalance.
\newblock In {\em International Conference on Discovery Science}, pages 159--174. Springer, 2020.

\bibitem[\protect\citeauthoryear{Iosifidis \bgroup \em et al.\egroup }{2019}]{iosifidis2019fairness}
Vasileios Iosifidis, Thi Ngoc~Han Tran, and Eirini Ntoutsi.
\newblock Fairness-enhancing interventions in stream classification.
\newblock In {\em DEXA}, pages 261--276. Springer, 2019.

\bibitem[\protect\citeauthoryear{Kallus and Zhou}{2018}]{kallus2018residual}
Nathan Kallus and Angela Zhou.
\newblock Residual unfairness in fair machine learning from prejudiced data.
\newblock In {\em ICML}, pages 2439--2448. PMLR, 2018.

\bibitem[\protect\citeauthoryear{K{\"a}rkk{\"a}inen and Joo}{2019}]{karkkainen2019fairface}
Kimmo K{\"a}rkk{\"a}inen and Jungseock Joo.
\newblock Fairface: Face attribute dataset for balanced race, gender, and age.
\newblock {\em arXiv:1908.04913}, 2019.

\bibitem[\protect\citeauthoryear{Koh \bgroup \em et al.\egroup }{2021}]{koh2021wilds}
Pang~Wei Koh, Shiori Sagawa, Henrik Marklund, Sang~Michael Xie, Marvin Zhang, Akshay Balsubramani, Weihua Hu, Michihiro Yasunaga, Richard~Lanas Phillips, Irena Gao, et~al.
\newblock Wilds: A benchmark of in-the-wild distribution shifts.
\newblock In {\em ICML}, pages 5637--5664. PMLR, 2021.

\bibitem[\protect\citeauthoryear{Kohavi and others}{1996}]{kohavi1996scaling}
Ron Kohavi et~al.
\newblock Scaling up the accuracy of naive-bayes classifiers: A decision-tree hybrid.
\newblock In {\em KDD}, volume~96, pages 202--207, 1996.

\bibitem[\protect\citeauthoryear{Kusner \bgroup \em et al.\egroup }{2017}]{kusner2017counterfactual}
Matt~J Kusner, Joshua Loftus, Chris Russell, and Ricardo Silva.
\newblock Counterfactual fairness.
\newblock {\em NeurIPS}, 30, 2017.

\bibitem[\protect\citeauthoryear{Li \bgroup \em et al.\egroup }{2022}]{li2022fair}
Shu Li, Jiong Yu, Xusheng Du, Yi~Lu, and Rui Qiu.
\newblock Fair outlier detection based on adversarial representation learning.
\newblock {\em Symmetry}, 14(2):347, 2022.

\bibitem[\protect\citeauthoryear{Li \bgroup \em et al.\egroup }{2024}]{li2024graph}
Yibo Li, Xiao Wang, Yujie Xing, Shaohua Fan, Ruijia Wang, Yaoqi Liu, and Chuan Shi.
\newblock Graph fairness learning under distribution shifts.
\newblock {\em arXiv preprint arXiv:2401.16784}, 2024.

\bibitem[\protect\citeauthoryear{Lin \bgroup \em et al.\egroup }{2023}]{lin2023adaptation}
Yujie Lin, Chen Zhao, Minglai Shao, Xujiang Zhao, and Haifeng Chen.
\newblock Adaptation speed analysis for fairness-aware causal models.
\newblock In {\em CIKM}, 2023.

\bibitem[\protect\citeauthoryear{Lin \bgroup \em et al.\egroup }{2024}]{lin2023pursuing}
Yujie Lin, Chen Zhao, Minglai Shao, Baoluo Meng, Xujiang Zhao, and Haifeng Chen.
\newblock Towards counterfactual fairness-aware domain generalization in changing environments.
\newblock {\em IJCAI}, 2024.

\bibitem[\protect\citeauthoryear{Liu \bgroup \em et al.\egroup }{2021}]{liu2021domain}
Xiaofeng Liu, Bo~Hu, Linghao Jin, Xu~Han, Fangxu Xing, Jinsong Ouyang, Jun Lu, Georges~EL Fakhri, and Jonghye Woo.
\newblock Domain generalization under conditional and label shifts via variational bayesian inference.
\newblock {\em arXiv:2107.10931}, 2021.

\bibitem[\protect\citeauthoryear{Mandal \bgroup \em et al.\egroup }{2020}]{mandal2020ensuring}
Debmalya Mandal, Samuel Deng, Suman Jana, Jeannette Wing, and Daniel~J Hsu.
\newblock Ensuring fairness beyond the training data.
\newblock {\em NeurIPS}, 2020.

\bibitem[\protect\citeauthoryear{Newman}{2015}]{newman2015semantic}
John Newman.
\newblock Semantic shift.
\newblock {\em The Routledge handbook of semantics}, pages 266--280, 2015.

\bibitem[\protect\citeauthoryear{Oh \bgroup \em et al.\egroup }{2022}]{oh2022learning}
Changdae Oh, Heeji Won, Junhyuk So, Taero Kim, Yewon Kim, Hosik Choi, and Kyungwoo Song.
\newblock Learning fair representation via distributional contrastive disentanglement.
\newblock In {\em KDD}, 2022.

\bibitem[\protect\citeauthoryear{Pearl}{2009}]{pearl2009causality}
Judea Pearl.
\newblock {\em Causality}.
\newblock Cambridge university press, 2009.

\bibitem[\protect\citeauthoryear{Pham \bgroup \em et al.\egroup }{2023}]{pham2023fairness}
Thai-Hoang Pham, Xueru Zhang, and Ping Zhang.
\newblock Fairness and accuracy under domain generalization.
\newblock {\em arXiv:2301.13323}, 2023.

\bibitem[\protect\citeauthoryear{Rezaei \bgroup \em et al.\egroup }{2020}]{rezaei2020fairness}
Ashkan Rezaei, Rizal Fathony, Omid Memarrast, and Brian Ziebart.
\newblock Fairness for robust log loss classification.
\newblock In {\em AAAI}, 2020.

\bibitem[\protect\citeauthoryear{Rezaei \bgroup \em et al.\egroup }{2021}]{rezaei2021robust}
Ashkan Rezaei, Anqi Liu, Omid Memarrast, and Brian~D Ziebart.
\newblock Robust fairness under covariate shift.
\newblock In {\em AAAI}, 2021.

\bibitem[\protect\citeauthoryear{Roh \bgroup \em et al.\egroup }{2023}]{roh2023improving}
Yuji Roh, Kangwook Lee, Steven~Euijong Whang, and Changho Suh.
\newblock Improving fair training under correlation shifts.
\newblock {\em ICML}, 2023.

\bibitem[\protect\citeauthoryear{Saerens \bgroup \em et al.\egroup }{2002}]{saerens2002adjusting}
Marco Saerens, Patrice Latinne, and Christine Decaestecker.
\newblock Adjusting the outputs of a classifier to new a priori probabilities: a simple procedure.
\newblock {\em Neural computation}, 14(1):21--41, 2002.

\bibitem[\protect\citeauthoryear{Sagawa \bgroup \em et al.\egroup }{2019}]{sagawa2019distributionally}
Shiori Sagawa, Pang~Wei Koh, Tatsunori~B Hashimoto, and Percy Liang.
\newblock Distributionally robust neural networks for group shifts: On the importance of regularization for worst-case generalization.
\newblock {\em arXiv:1911.08731}, 2019.

\bibitem[\protect\citeauthoryear{Schrouff \bgroup \em et al.\egroup }{2022}]{schrouff2022diagnosing}
Jessica Schrouff, Natalie Harris, Sanmi Koyejo, Ibrahim~M Alabdulmohsin, Eva Schnider, Krista Opsahl-Ong, Alexander Brown, Subhrajit Roy, Diana Mincu, Christina Chen, et~al.
\newblock Diagnosing failures of fairness transfer across distribution shift in real-world medical settings.
\newblock {\em NeurIPS}, 35:19304--19318, 2022.

\bibitem[\protect\citeauthoryear{Schumann \bgroup \em et al.\egroup }{2019}]{schumann2019transfer}
Candice Schumann, Xuezhi Wang, Alex Beutel, Jilin Chen, Hai Qian, and Ed~H Chi.
\newblock Transfer of machine learning fairness across domains.
\newblock {\em arXiv:1906.09688}, 2019.

\bibitem[\protect\citeauthoryear{Shekhar \bgroup \em et al.\egroup }{2021}]{shekhar2021fairod}
Shubhranshu Shekhar, Neil Shah, and Leman Akoglu.
\newblock Fairod: Fairness-aware outlier detection.
\newblock In {\em AIES}, pages 210--220, 2021.

\bibitem[\protect\citeauthoryear{Shimodaira}{2000}]{shimodaira2000improving}
Hidetoshi Shimodaira.
\newblock Improving predictive inference under covariate shift by weighting the log-likelihood function.
\newblock {\em Journal of statistical planning and inference}, 2000.

\bibitem[\protect\citeauthoryear{Shpitser and Pearl}{2008}]{shpitser2008complete}
Ilya Shpitser and Judea Pearl.
\newblock Complete identification methods for the causal hierarchy.
\newblock {\em JMLR}, 2008.

\bibitem[\protect\citeauthoryear{Singh \bgroup \em et al.\egroup }{2021}]{singh2021fairness}
Harvineet Singh, Rina Singh, Vishwali Mhasawade, and Rumi Chunara.
\newblock Fairness violations and mitigation under covariate shift.
\newblock In {\em FAccT}, 2021.

\bibitem[\protect\citeauthoryear{Song \bgroup \em et al.\egroup }{2021}]{song2021deep}
Hanyu Song, Peizhao Li, and Hongfu Liu.
\newblock Deep clustering based fair outlier detection.
\newblock In {\em SIGKDD}, pages 1481--1489, 2021.

\bibitem[\protect\citeauthoryear{Taskesen \bgroup \em et al.\egroup }{2020}]{taskesen2020distributionally}
Bahar Taskesen, Viet~Anh Nguyen, Daniel Kuhn, and Jose Blanchet.
\newblock A distributionally robust approach to fair classification.
\newblock {\em arXiv:2007.09530}, 2020.

\bibitem[\protect\citeauthoryear{Thomas \bgroup \em et al.\egroup }{2019}]{thomas2019preventing}
Philip~S Thomas, Bruno Castro~da Silva, Andrew~G Barto, Stephen Giguere, Yuriy Brun, and Emma Brunskill.
\newblock Preventing undesirable behavior of intelligent machines.
\newblock {\em Science}, 366(6468):999--1004, 2019.

\bibitem[\protect\citeauthoryear{Wang \bgroup \em et al.\egroup }{2003}]{wang2003mining}
Ke~Wang, Senqiang Zhou, Chee~Ada Fu, and Jeffrey~Xu Yu.
\newblock Mining changes of classification by correspondence tracing.
\newblock In {\em Proceedings of the 2003 SIAM International Conference on Data Mining}. SIAM, 2003.

\bibitem[\protect\citeauthoryear{Widmer and Kubat}{1996}]{conceptshift}
Gerhard Widmer and Miroslav Kubat.
\newblock Learning in the presence of concept drift and hidden contexts.
\newblock {\em Machine learning}, 23:69--101, 1996.

\bibitem[\protect\citeauthoryear{Yamazaki \bgroup \em et al.\egroup }{2007}]{yamazaki2007asymptotic}
Keisuke Yamazaki, Motoaki Kawanabe, Sumio Watanabe, Masashi Sugiyama, and Klaus-Robert M{\"u}ller.
\newblock Asymptotic bayesian generalization error when training and test distributions are different.
\newblock In {\em ICML}, pages 1079--1086, 2007.

\bibitem[\protect\citeauthoryear{Yao and Liu}{2023}]{yao2023understanding}
Yuanshun Yao and Yang Liu.
\newblock Understanding unfairness via training concept influence.
\newblock {\em arXiv:2306.17828}, 2023.

\bibitem[\protect\citeauthoryear{Yoon \bgroup \em et al.\egroup }{2020}]{yoon2020joint}
Taeho Yoon, Jaewook Lee, and Woojin Lee.
\newblock Joint transfer of model knowledge and fairness over domains using wasserstein distance.
\newblock {\em IEEE Access}, 8:123783--123798, 2020.

\bibitem[\protect\citeauthoryear{Zemel \bgroup \em et al.\egroup }{2013}]{zemel2013learning}
Rich Zemel, Yu~Wu, Kevin Swersky, Toni Pitassi, and Cynthia Dwork.
\newblock Learning fair representations.
\newblock In {\em ICML}. PMLR, 2013.

\bibitem[\protect\citeauthoryear{Zhang \bgroup \em et al.\egroup }{2017}]{zhang2017age}
Zhifei Zhang, Yang Song, and Hairong Qi.
\newblock Age progression/regression by conditional adversarial autoencoder.
\newblock In {\em CVPR}, pages 5810--5818, 2017.

\bibitem[\protect\citeauthoryear{Zhao \bgroup \em et al.\egroup }{2021}]{zhao2021fairness}
Chen Zhao, Feng Chen, and Bhavani Thuraisingham.
\newblock Fairness-aware online meta-learning.
\newblock In {\em SIGKDD}, pages 2294--2304, 2021.

\bibitem[\protect\citeauthoryear{Zhao \bgroup \em et al.\egroup }{2022}]{zhao2022adaptive}
Chen Zhao, Feng Mi, Xintao Wu, Kai Jiang, Latifur Khan, and Feng Chen.
\newblock Adaptive fairness-aware online meta-learning for changing environments.
\newblock In {\em SIGKDD}, pages 2565--2575, 2022.

\bibitem[\protect\citeauthoryear{Zhao \bgroup \em et al.\egroup }{2023a}]{zhao2023fairness}
Chen Zhao, Kai Jiang, Xintao Wu, Haoliang Wang, Latifur Khan, Christan Grant, and Feng Chen.
\newblock Fairness-aware domain generalization under covariate and dependence shifts.
\newblock {\em arXiv:2311.13816}, 2023.

\bibitem[\protect\citeauthoryear{Zhao \bgroup \em et al.\egroup }{2023b}]{zhao2023towards}
Chen Zhao, Feng Mi, Xintao Wu, Kai Jiang, Latifur Khan, Christan Grant, and Feng Chen.
\newblock Towards fair disentangled online learning for changing environments.
\newblock In {\em SIGKDD}, pages 3480--3491, 2023.

\end{thebibliography}

\end{document}